\documentclass[conference]{IEEEtran}

\IEEEoverridecommandlockouts

\usepackage{color}
\usepackage{pifont}
\usepackage{xcolor}
\usepackage{xspace}
\usepackage{acronym}
\usepackage{graphicx}
\usepackage{multicol}
\usepackage{multirow}
\usepackage{textcomp}
\usepackage{hyperref} 
\usepackage{algorithmic}
\usepackage{amsmath,amssymb,amsfonts}

\acrodef{BoW}{Bag of Words}
\acrodef{VO}{Visual Odometry}
\acrodef{CV}{Computer Vision}
\acrodef{UWB}{Ultra-wideband}
\acrodef{BA}{Bundle Adjustment}
\acrodef{AR}{Augmented Reality}
\acrodef{DoF}{Degree of Freedom}
\acrodef{KLT}{Kanade-Lucas-Tomasi}
\acrodef{LSD}{Line Segment Detector}
\acrodef{OGM}{Occupancy Grid Mapping}
\acrodef{RGB-D}{Red Green Blue-Depth}
\acrodef{EKF}{Extended Kalman Filter}
\acrodef{DSO}{Direct Sparse Odometry}
\acrodef{PGO}{Pose-Graph Optimization}
\acrodef{LSTM}{Long Short-Term Memory}
\acrodef{UAV}{Unmanned Aerial Vehicle}
\acrodef{UGV}{Unmanned Ground Vehicle}
\acrodef{VAE}{Variational Auto-Encoder}
\acrodef{RNN}{Recurrent Neural Network}
\acrodef{GPU}{Graphics Processing Unit}
\acrodef{YOLO2}{You Only Look Once v2.0}
\acrodef{IMU}{Inertial Measurement Unit}
\acrodef{NRCC}{Non-Rigid Context Culling}
\acrodef{RANSAC}{Random Sample Consensus}
\acrodef{SURF}{Speeded Up Robust Features}
\acrodef{SVO}{Semi-direct Visual Odometry}
\acrodef{DTAM}{Dense Tracking and Mapping}
\acrodef{CNN}{Convolutional Neural Network}
\acrodef{GMS}{Grid-based Motion Statistics}
\acrodef{EPnP}{Efficient Perspective-n-Point}
\acrodef{PTAM}{Parallel Tracking and Mapping}
\acrodef{ORB}{Oriented FAST and Rotated BRIEF}
\acrodef{SIFT}{Scale-Invariant Feature Transform}
\acrodef{SLAM}{Simultaneous Localization and Mapping}
\acrodef{RCNN}{Recurrent Convolutional Neural Network}
\acrodef{FAST}{Features from Accelerated Segment Test}
\acrodef{VSLAM}{Visual Simultaneous Localization and Mapping}
\acrodef{BRIEF}{Binary Robust Independent Elementary Features}

\newcommand{\cmark}{\ding{51}}

\newcommand{\etc}{\textit{etc. }}
\newcommand{\eg}{\textit{e.g., }}
\newcommand{\ie}{\textit{i.e., }}
\newcommand{\etal}{\textit{et al. }}

\def\BibTeX{{\rm B\kern-.05em{\sc i\kern-.025em b}\kern-.08em
    T\kern-.1667em\lower.7ex\hbox{E}\kern-.125emX}}
\begin{document}

\title{Visual SLAM: What are the Current Trends and What to Expect?\\

}

\author{
    \IEEEauthorblockN{
        Ali Tourani\IEEEauthorrefmark{1},
        Hriday Bavle\IEEEauthorrefmark{2},
        Jose-Luis Sanchez-Lopez\IEEEauthorrefmark{3},
        and Holger Voos\IEEEauthorrefmark{4}
    } \\
    \IEEEauthorblockA{
        \IEEEauthorrefmark{1}\IEEEauthorrefmark{2}\IEEEauthorrefmark{3}\IEEEauthorrefmark{4}
        University of Luxembourg, Interdisciplinary Centre for Security, Reliability, and Trust (SnT), \\ L-1855 Luxembourg, Luxembourg \\
        \IEEEauthorrefmark{4}University of Luxembourg, Department of Engineering, L-1359 Luxembourg, Luxembourg \\
        Email:
        \IEEEauthorrefmark{1}ali.tourani@uni.lu,
        \IEEEauthorrefmark{2}hriday.bavle@uni.lu,
        \IEEEauthorrefmark{3}joseluis.sanchezlopez@uni.lu,
        \IEEEauthorrefmark{4}holger.voos@uni.lu
    }
}



\maketitle

\begin{abstract}
Vision-based sensors have shown significant performance, accuracy, and efficiency gain in \ac{SLAM} systems in recent years.
In this regard, \ac{VSLAM} methods refer to the SLAM approaches that employ cameras for pose estimation and map generation.
We can see many research works that demonstrated \acp{VSLAM} can outperform traditional methods, which rely only on a particular sensor, such as a Lidar, even with lower costs.
\ac{VSLAM} approaches utilize different camera types (\eg monocular, stereo, and RGB-D), have been tested on various datasets (\eg KITTI, TUM RGB-D, and EuRoC) and in dissimilar environments (\eg indoors and outdoors), and employ multiple algorithms and methodologies to have a better understanding of the environment.
The mentioned variations have made this topic popular for researchers and resulted in a wide range of \acp{VSLAM} methodologies.
In this regard, the primary intent of this survey is to present the recent advances in \ac{VSLAM} systems, along with discussing the existing challenges and trends.
We have given an in-depth literature survey of forty-five impactful papers published in the domain of \acp{VSLAM}.
We have classified these manuscripts by different characteristics, including the novelty domain, objectives, employed algorithms, and semantic level.
We also discuss the current trends and future directions that may help researchers investigate them.
\end{abstract}


\section{Introduction}
\label{sec_intro}

\begin{figure*}[t!]
    \centering
    \caption{The flowchart of a standard visual \ac{SLAM} approach. Regarding the direct/indirect methodology utilized, the functionality of some of these modules may change or ignored.}
    \label{fig_flowchart}
    \includegraphics[width=0.9\linewidth]{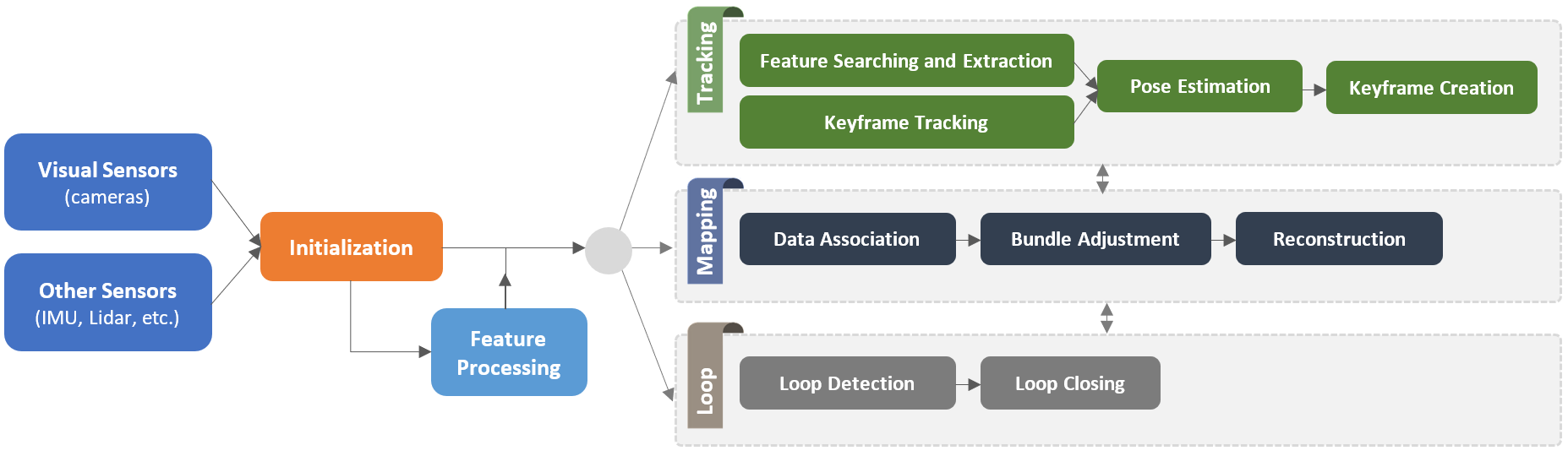}
\end{figure*}

\acf{SLAM} refers to the process of estimating an unknown environment's map while monitoring the location of an \textit{agent} at the same time \cite{khairuddin2015review}.
Here, the \textit{agent} can be a domestic robot \cite{vallivaara2011magnetic}, an autonomous vehicle \cite{zou2021comparative}, a planetary rover \cite{geromichalos2020slam}, even an \ac{UAV} \cite{yang2018monocular, li2016real} or an \ac{UGV} \cite{liu2018real}.
In situations where a prior map of the environment is unavailable or the robot's location is unknown, \ac{SLAM} can be utilized to cover a wide range of applications.
In this regard, and considering the ever-growing applications of robotics, \ac{SLAM} has gained huge attention among industry and research community members in recent years \cite{gupta2022simultaneous, cadena2016past}.

\ac{SLAM} systems may use various sensors to collect data from the environment, including laser-based sensors, acoustic, and vision sensors \cite{zaffar2018sensors}.
The vision sensors category covers any variety of visual data detectors, including monocular, stereo, event-based, omnidirectional, and \ac{RGB-D} cameras.
A robot equipped with a vision sensor uses the visual data provided by cameras to estimate the position and orientation of the robot with respect to its surroundings \cite{gao2021introduction}.
The process of using vision sensors to perform \ac{SLAM} is particularly called \acf{VSLAM}.
Utilizing visual data in \ac{SLAM} applications has the advantages of cheaper hardware requirement, more straightforward object detection and tracking, and the ability to provide rich visual and semantic information \cite{filipenko2018comparison}.
The captured images (or video frames) can also be used for vision-based applications, including semantic segmentation and object detection, as they store a wealth of data for processing.
The mentioned characteristics have recently made \ac{VSLAM} a trending topic in robotics and prompted robotics and \ac{CV} experts to perform considerable studies and investigations in the last decades.
Consequently, \ac{VSLAM} can be found in various types of applications where it is essential to reconstruct the 3D model of the environment, such as autonomous, \ac{AR}, and service robots \cite{yeh20183d}.

As a general benchmark introduced by \cite{klein2007parallel} to tackle high computational cost, \ac{SLAM} approaches mainly contain two introductory threads to be executed in parallel, known as \textit{tracking} and \textit{mapping}.
Hereby, a fundamental classification of the algorithms used in \ac{VSLAM} is how researchers employ distinct methods and strategies in each thread.
The mentioned solutions look differently at \ac{SLAM} systems based on the type of data they use, making them dividable into two categories: \textit{Direct} and \textit{Indirect (feature-based)} methods \cite{duan2020deep}.
\textit{Indirect} methods extract feature points (\ie keypoints) obtained from textures by processing the scene and keep track of them by matching their descriptors in sequential frames.
Despite the computationally expensive performance of feature extraction and matching stages, these methods are precise and robust against photo-metric changes in frame intensities.
\textit{Direct} algorithms, on the other hand, estimate camera motions directly from pixel-level data and build an optimization problem to minimize the photo-metric error.
By relying on photogrammetry, these methods utilize all camera output pixels and track their replacement in sequential frames regarding their constrained aspects, such as brightness and color.
These characteristic enable \textit{direct} approaches to model more information from images than \textit{indirect} techniques and enable a higher-accuracy $3D$ reconstruction.
However, while direct methods work better in texture-less environments and do not require more computation for feature extraction, they often face large-scale optimization problem \cite{outahar2021direct}.
The pros and cons of each approach encouraged researchers to think about developing \textit{Hybrid} solutions, where a combination of both approaches are considered.
\textit{Hybrid} methods commonly integrate the detection stage of \textit{indirect} and \textit{direct}, in which one initializes and corrects the other.

Additionally, as \acp{VSLAM} mainly include a \ac{VO} front-end to locally estimate the path of the camera and a \ac{SLAM} back-end to optimize the created map, the variety of modules used in each category results in implementation variations.
\ac{VO} provides preliminary estimation of the location and pose of the robot based on local consistencies and sent to the back-end for optimization.
Thus, the primary distinction between \ac{VSLAM} and \ac{VO} is whether or not to take into account the global consistency of the map and the predicted trajectory.
Several state-of-the-art \ac{VSLAM} applications also include two additional modules: loop closure detection and mapping \cite{duan2020deep}.
They are responsible for recognizing previously visited locations for more precise tracking and map reconstruction based on the camera pose.

To summarize, \figurename{\ref{fig_flowchart}} shows the overall architecture of a standard \ac{VSLAM} approach.
Accordingly, the system's inputs may also integrate with other sensor data, such as \ac{IMU} and Lidar, to provide more information rather than visual data.
Moreover, regarding the \textit{direct} or \textit{indirect} methodology used in a \ac{VSLAM} pipeline, the functionality of the visual features processing module might be changed or ignored.
For instance, the "Feature Processing" stage is only employed in \textit{indirect} approaches.
Another factor is utilizing some particular modules such as loop closing detection and bundle adjustment for improved execution.

This paper surveys forty-five \ac{VSLAM} papers and classifies them into various categories according to diverse aspects.
We hope our work will present a reference for the robotics community researchers working to improve \ac{VSLAM} techniques.
The rest of the paper is organized as follows:
Section~\ref{sec_evolution} reviews the evolutionary stages in \ac{VSLAM} methods that lead to the currently existing systems.
We introduce and discuss other published surveys in \ac{VSLAM} domain in Section~\ref{sec:related_survey}.
An abstract-level of various \ac{VSLAM} modules are presented in Section~\ref{sec:literature} and a classification of state-of-the-arts based on the main contributions are available in Section~\ref{sec:focus}.
We will then talk about unresolved challenges and potential trends in this field in Section~\ref{sec:discussion}.
The paper finally concludes in Section~\ref{sec:conclusion}.
\section{Evolution of Visual SLAM Algorithms}
\label{sec_evolution}

\begin{figure*}[t!]
    \centering
    \caption{Milestones of highly impactful Visual SLAM approaches.}
    \label{fig:timeline}
    \includegraphics[width=0.9\linewidth]{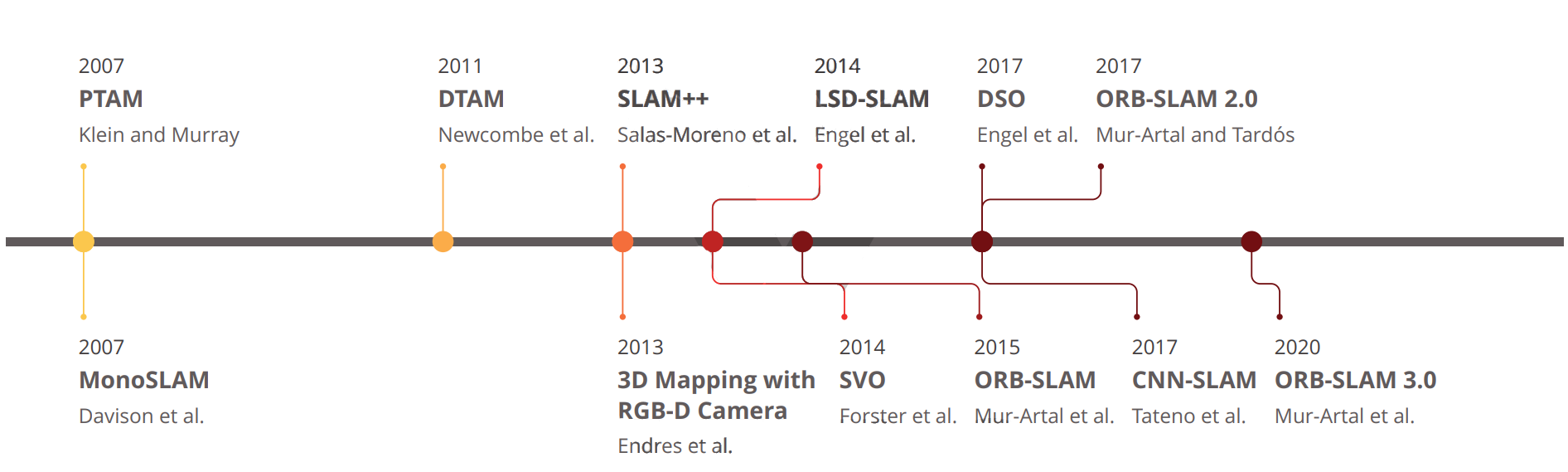}
\end{figure*}

\ac{VSLAM} systems have matured over the past years, and several frameworks have played a vital role in this development process.
To provide a general picture, \figurename{\ref{fig:timeline}} illustrates the milestones of the widely-referred \ac{VSLAM} approaches that impacted the community and have been used as baselines for other frameworks.

Accordingly, the first endeavor in the literature to implement a real-time monocular \ac{VSLAM} system was developed by Davison \etal in $2007$ where they introduced a framework titled \textit{Mono-SLAM} \cite{davison2007monoslam}.
Their indirect framework could estimate the camera motion and 3D elements found in the world using the \ac{EKF} algorithm \cite{ribeiro2004kalman}.
\textit{Mono-SLAM} began the primary action in the \ac{VSLAM} domain despite the lack of global optimization and loop closure detection modules.
However, the maps reconstructed by this method only include landmarks and do not offer further detailed information about the area.
Klein \etal in \cite{klein2007parallel} proposed \ac{PTAM} in the same year, in which they divided the entire \ac{VSLAM} system into two primary threads: \textit{tracking} and \textit{mapping}.
This multi-threading baseline was approved by many subsequent works in later works, which will be discussed in this paper.
The main idea in their approach was to reduce the computational cost and apply parallel processing to achieve real-time performance.
While the \textit{tracking} thread estimates camera motion in real-time, \textit{mapping} predicts the $3D$ positions of feature points.
\ac{PTAM} was also the first approach to utilize \ac{BA} for jointly optimizing the camera poses and the created $3D$ map.
It uses \ac{FAST} \cite{viswanathan2009features} corner detector algorithm for key-points matching and tracking.
Despite better performance than \textit{Mono-SLAM}, the algorithm has complex design and requires user input in the first stage. 
A direct approach for measuring depth values and motion parameters for map construction was \ac{DTAM} introduced by Newcombe \etal \cite{newcombe2011dtam} in 2011.
\ac{DTAM} was a real-time framework equipped with \textit{dense mapping} and \textit{dense tracking} modules and could determine camera poses by aligning the entire frames with a given depth map.
To construct the environment map, the mentioned stages estimate the depth of the scene and the motion parameters, respectively.
Although \ac{DTAM} can provide a detailed presentation of the map, it demands high computational cost to perform in real-time.
As another indirect approach in the domain of 3D mapping and pixel-based optimization, Endres \etal \cite{endres20133} in 2013 proposed a method that could work with \ac{RGB-D} cameras.
Their method performs in real-time and is focused on low-cost embedded systems and small robots, but it cannot produce accurate results in featureless or challenging scenarios.
At the same year, Salas-Moreno \etal \cite{salas2013slam++} proposed one of the first endeavors in utilizing semantic information in a real-time \ac{SLAM} framework, titled SLAM++.
Their system employs \ac{RGB-D} sensor outputs and performs 3D camera pose estimation and tracking to shape a pose graph.
A pose graph is a graph in which the nodes represent pose estimates and are connected by edges representing the relative poses between nodes with measurement uncertainty \cite{mendes2016icp}.
The predicted poses will then be optimized by merging the relative 3D poses obtained from semantic objects in the scene.

With ripening the baseline of \ac{VSLAM}, researchers focused on improving the performance and precision of these systems.
In this regard, Forster \etal in 2014 proposed a hybrid \ac{VO} approach known as \ac{SVO} \cite{forster2014svo} as a part of \ac{VSLAM} architectures.
Their method could merge feature-based and direct approaches to perform sensors' motion estimation and mapping tasks.
\ac{SVO} could work with both monocular and stereo cameras and was equipped with a pose refinement module to minimize re-projection errors.
However, the main drawbacks of \ac{SVO} are employing a short-term data association and the inability to perform loop closure detection and global optimization.
LSD-SLAM \cite{engel2014lsd} is another influential \ac{VSLAM} method introduced by Engel \etal in 2014 and contains \textit{tracking}, \textit{depth map estimation}, and \textit{map optimization} threads.
The method could reconstruct large-scale maps using its pose-graph estimation module and was equipped with global optimization and loop closure detection feat.
The weakness of LSD-SLAM is its challenging initialization stage that requires all points in a plane, making it a computationally intensive approach.
Mur-Artal \etal introduced two accurate indirect \ac{VSLAM} approaches that have attracted the attention of many researchers so far: ORB-SLAM \cite{mur2015orb} and ORB-SLAM 2.0 \cite{mur2017orb}.
These methods can accomplish localization and mapping in well-textured sequences and perform high-performance position recognition using \ac{ORB} features.
The first version of ORB-SLAM is able to compute both the camera position and the environment's structure using the keyframes collected from camera locations.
The second version is the extension to ORB-SLAM with three parallel threads, including \textit{tracking} for finding feature correspondences, \textit{local mapping} for map management operations, and \textit{loop closing} for detecting new loops and correcting the drift error.
Although ORB-SLAM 2.0 can work with both monocular and stereo camera setups, it cannot be used for autonomous navigation due to reconstructing maps with unknown scales.
Another drawback of this approach is its inability to work in texture-less areas or environments with repetitive patterns.
The most recent version of this framework, named ORB-SLAM 3.0, was proposed in 2021 \cite{campos2021orb}.
It works with various camera types, such as monocular, \ac{RGB-D} and stereo-vision, and provides improved pose estimation outputs.

In recent years and with the significant influences of deep learning in various domains, deep neural network-based approaches could resolve many issues by providing higher recognition and matching rates.
Similarly, replacing hand-crafted with learned features in \ac{VSLAM} is one of the solutions suggested by many recent deep learning-based methods.
In this regard, Tateno \etal presented an approach based on \acp{CNN} that processes the input frames for camera pose estimation and uses keyframes for depth prediction, anointed CNN-SLAM \cite{tateno2017cnn}.
Segmenting camera frames into smaller sections to provide better understanding of the environment is one of the ideas in CNN-SLAM to provide parallel processing and real-time performance.
As a different methodology, Engel \etal also introduced a new trend in direct \ac{VSLAM} algorithms titled \ac{DSO} \cite{engel2017direct} that merges a direct approach and a sparse reconstruction to extract the highest intensity points in image blocks.
By tracking sparse sets of pixels, it considers the image formation parameters and uses an indirect tracking method.
It should be noted that \ac{DSO} can only provide perfect accuracy if the photo-metrically calibrated cameras are used and fails to achieve high-accuracy results using regular cameras.

To recap, milestones in the \ac{VSLAM} systems evolution process reveal that recent approaches focus on the parallel execution of multiple dedicated modules.
These modules shaped general-purpose techniques and frameworks compatible with a broad range of sensors and environments.
The mentioned characteristic enables them to be executable in real-time and be more flexible in terms of performance improvement.
\section{Related Surveys}
\label{sec:related_survey}

There are various survey papers available in the domain of \ac{VSLAM} that present a general review of the different existing approaches.
Each of these papers review the major advantages and disadvantages of employing \ac{VSLAM} approaches.
Macario Barros \etal \cite{macario2022comprehensive} divided approaches in three different classes: visual-only (monocular), visual-inertial (stereo), and \ac{RGB-D}.
They also proposed various criteria for simplifying analyzing \ac{VSLAM} algorithms.
However, they did not include other vision sensors, such as event camera-based ones, which we will discuss later in \ref{sec:sensors}.
Chen \etal \cite{chen2022overview} reviewed a wide range of traditional and semantic \ac{VSLAM} publications.
They divided the \ac{SLAM} development era into \textit{classical}, \textit{algorithmic-analysis}, and \textit{robust-perception} stages and introduced hot issues there.
They also summarized classical frameworks that employ direct/indirect methodologies and investigated the impact of deep learning algorithms in semantic segmentation.
Although their work provides a comprehensive study of the advanced solutions in this domain, the classification of approaches is only restricted to the \textit{feature types} employed in feature-based \acp{VSLAM}.
Jia \etal \cite{jia2019survey} surveyed numerous manuscripts and presented a brief comparison between graph optimization-based methods and deep learning-equipped approaches.
Despite presenting a proper comparison, their discussion cannot be generalized due to reviewing a limited number of papers.
In another work, Abaspur Kazerouni \etal \cite{kazerouni2022survey} covered various \ac{VSLAM} methods, utilized sensory equipment, datasets, and modules and simulated several indirect approaches for comparison and analysis.
They contribute only to the feature-based algorithms \textit{-\eg HOG, \ac{SIFT}, and \ac{SURF}} and deep learning-based solutions.
Bavle \etal \cite{bavle2021slam} analyzed the situational awareness aspects in various \ac{SLAM} and \ac{VSLAM} applications and discussed their missing points.
They could conclude that operating the lacking situational awareness features could enhance the performance of the current research works.

Other surveys studied the latest \ac{VSLAM} approaches focused on a particular topic or trend.
For instance, Duan \etal \cite{duan2020deep} investigated the progress of deep learning in visual \ac{SLAM} systems for transportation robotics.
The authors summarized the advantages and drawbacks of utilizing various deep learning-based methods in \ac{VO} and loop closure detection tasks in their paper.
The significant advantage of using deep learning approaches in \acp{VSLAM} is the accurate feature extraction in pose estimation and the overall performance calculation.
In another work in the same field, Arshad and Kim \cite{arshad2021role} focused on the impact of deep learning algorithms in loop closure detection using visual data.
They reviewed various \ac{VSLAM} papers and analyzed the long-term autonomy of robots in different conditions.
Singandhupe and La \cite{singandhupe2019review} reviewed the impact of \ac{VO} and \ac{VSLAM} in driverless vehicles.
They collected approaches that have been evaluated on the KITTI dataset, enabling them to have a brief description of the advantages and demerits of each system.
Cheng \etal \cite{chen2022overview} in a similar manuscript reviewed the \ac{VSLAM}-based autonomous driving systems and raised the future development trends of such systems.
Some other researchers surveyed VSLAM works with the ability to work in real-world conditions.
For instance, Saputra \etal \cite{saputra2019vslam} targeted the variations of \ac{VSLAM} techniques operating in dynamic and rough environments and discussed the reconstruction, segmentation, tracking, and parallel execution of threads problems.

Regarding the mentioned surveys, the current survey has particularities that set it apart from other surveys presented so far and provides a comprehensive review of the \ac{VSLAM} systems presented in different venues.
In this regard, the major contributions of this survey compared to other available \ac{VSLAM} surveys are:

\begin{itemize}
    \item Categorizing various recent \ac{VSLAM} publications regarding the main contributions, criteria, and objectives of researchers in proposing new solutions,
    \item Analyzing the current trends of \ac{VSLAM} systems by profoundly investigating different approaches regarding dissimilar aspects,
    \item Introducing the potential contributions of \ac{VSLAM} for researchers
\end{itemize}
\section{VSLAM Setup Criteria}
\label{sec:literature}

Considering various \ac{VSLAM} approaches, we can classify different setups and configurations available into categories mentioned below:

\subsection{Sensors and Data Acquisition}
\label{sec:sensors}

The early-stage implementation of a \ac{VSLAM} algorithm introduced by Davison \etal \cite{davison2007monoslam} was equipped with a monocular camera for trajectory recovery.
Monocular cameras are the most common vision sensors for a wide range of tasks, such as object detection and tracking \cite{he2020review}.
Stereo cameras, on the other hand, contain two or more image sensors, enabling them to perceive depth in the captured images, which leads to more accurate performance in \ac{VSLAM} applications.
These camera setups are cost-efficient and provide informative perception for higher accuracy demands.
\ac{RGB-D} cameras are other variations of visual sensors used in \acp{VSLAM} and supply both the depth and colors in the scene.
The mentioned vision sensors can provide rich information about the environment in straightforward circumstances -\textit{\eg proper lighting and motion speed}- but they often struggle to cope with conditions where the illumination is low or the dynamic range in the scene is high.

In recent years, event cameras have also been used in various \ac{VSLAM} applications.
These low latency bio-inspired vision sensors generate pixel-level brightness changes instead of standard intensity frames when a motion is detected, leading to a high dynamic range output with no motion blur impact \cite{gallego2020event}.
In contrast with standard cameras, event-based sensors supply trustworthy visual information during high-speed motions and wide-range dynamic scenarios but fail to provide sufficient information when the motion rate is low.
Although event cameras can outperform standard visual sensors in severe illumination and dynamic range conditions, they mainly generate unsynchronized information about the environment.
This makes traditional vision algorithms unable to process the outputs of these sensors \cite{jiao2021comparing}.
Additionally, using the spatio-temporal windows of events along with the data obtained from other sensors can provide rich pose estimation and tracking information.

Moreover, some approaches use a multi-camera setup to counter the common issues of working in a real-world environment and improve localization precision.
Utilizing multiple visual sensors aid in situations where complicated problems such as occlusion, camouflage, sensor failure, or sparsity of trackable texture occurs by providing cameras with overlapping field of views.
Although multi-camera setups can resolve some data acquisition issues, camera-only \acp{VSLAM} may face various challenges such as motion blur when encountering fast-moving objects, features mismatching in low or severe illumination, dynamic object ignorance in scenarios with high pace changes, \etc
Hence, some \ac{VSLAM} applications may equip with multiple sensors alongside cameras.
Fusing the events and standard frames \cite{vidal2018ultimate} or integrating other sensors such as Lidars \cite{xu2019occupancy} and \acp{IMU} to \ac{VSLAM} are some of the existing solutions.
\subsection{Target Environments}
\label{sec:env}

As a strong presumption in many traditional \ac{VSLAM} practices, the robot works in a static world with no sudden or unanticipated changes.
Consequently, although many systems could demonstrate successful application in specific settings, some unexpected changes in the environment (\eg the existence of moving objects) are likely to cause complications for the system and degrade the state estimation quality to a large extent.
Systems that work in dynamic environments usually employ algorithms such as Optical Flow or \ac{RANSAC} \cite{fischler1981random} to detect movements in the scene, classify the moving objects as outliers, and skip them while reconstructing the map.
Such systems utilize either \textit{geometry/semantic} information or try to improve the localization scheme by combining the results of these two \cite{cui2019sof}.

Additionally, we can classify different environments into \textit{indoor} and \textit{outdoor} categories as a general taxonomy.
An \textit{outdoor} environment can be an \textit{urban} area with structural landmarks and massive motion changes such as buildings and road textures, or an \textit{off-road} zone with a weak motion state such as moving clouds and vegetation, the texture of the sand, \etc
As a result of this, the amount of trackable points in \textit{off-road} environments is less than the \textit{urban} areas, which increases the risk of localization and loop closure detection failure.
\textit{Indoor} environments, on the other hand, contain scenes with entirely different global spatial properties, such as corridors, walls, and rooms.
We can anticipate that while a \ac{VSLAM} system might work well in one of the mentioned zones, it might not show the same performance in other environments.
\subsection{Visual Features Processing}
\label{sec:recog}

As discussed in Section \ref{sec_intro}, detecting visual features and utilizing feature descriptors information for pose estimation is an inevitable stage of indirect \ac{VSLAM} methodologies.
These approaches employ various feature extraction algorithms to understand the environment better and track the feature points in consecutive frames.
Feature extraction stage contains a wide range of algorithms, including \ac{SIFT}~\cite{lowe2004distinctive}, \ac{SURF}~\cite{bay2006surf}, \ac{FAST}~\cite{viswanathan2009features}, \ac{BRIEF}~\cite{calonder2010brief}, \ac{ORB}~\cite{rublee2011orb}, \etc
Among them, \ac{ORB} features have the advantage of fast extraction and matching without losing huge accuracy compared to \ac{SIFT} and \ac{SURF} \cite{karami2017image}.

The problem with some of the mentioned methods is that they cannot effectively adapt to various complex and unforeseen situations.
Thus, many researchers employed \acp{CNN} to extract deep-seated features of images for various stages, including \ac{VO}, pose estimation, and loop closure detection.
These techniques may represent supervised or unsupervised frameworks according to the functionality of the methods.
\subsection{System Evaluation}
\label{sec:test}

While some of the \ac{VSLAM} approaches, especially those with the capability of working in dynamic and challenging environments are tested on robots in real-world conditions, many research works have used publicly available datasets to demonstrate their applicability.
In this regard, the \textit{RAWSEEDS Dataset} by Bonarini \etal \cite{bonarini2006rawseeds} is a well-known multi-sensor benchmarking tool, containing indoor, outdoor, and mixed robot trajectories with ground-truth data.
It is one of the oldest publicly available benchmarking tools for robotic and \ac{SLAM} purposes.
\textit{Scenenet RGB-D} by McCormac \etal \cite{mccormac2017scenenet} is another favored dataset for scene understanding problems, such as semantic segmentation and object detection, containing five million large-scale rendered \ac{RGB-D} images.
The dataset also contains pixel-perfect ground-truth labels and exact camera poses and depth data, making it a potent tool for \ac{VSLAM} applications.
Many recent works in the domain of \ac{VSLAM} and \ac{VO} have tested their approaches on the \textit{TUM RGB-D} dataset \cite{sturm2012benchmark}.
The mentioned dataset and benchmarking tool contains color and depth images captured by a Microsoft Kinect sensor and their corresponding ground-truth sensors trajectories.
Also, \textit{NTU VIRAL} by Nguyen \etal \cite{nguyen2021ntuviral} is a dataset collected by a \ac{UAV} equipped with 3D lidars, cameras, \acp{IMU}, and multiple \acp{UWB}.
The dataset contains indoor and outdoor instances and targeted for evaluating autonomous driving and aerial operation performances.

Moreover, \textit{EuRoC MAV} by Burri \etal \cite{burri2016euroc} is another popular dataset containing images captured by a stereo camera, along with synchronized \ac{IMU} measurements and motion ground-truth.
Collected data in \textit{EuRoC MAV} are classified into easy, medium, and difficult categories according to the surrounding conditions.
\textit{OpenLORIS-Scene} by Shi \etal \cite{shi2020we} is another publicly available dataset for \ac{VSLAM} works, containing a wide range of data collected by a wheeled robot equipped with various sensors.
It provides proper data for monocular and \ac{RGB-D} algorithms, along with odometry data from wheel encoders.
As a more general purpose dataset used in \ac{VSLAM} applications, \textit{KITTI} \cite{geiger2012we} is popular collection of data, captured by two high-resolution RGB and grayscale video cameras on a moving vehicle.
\textit{KITTI} provides accurate ground-truth using GPS and laser sensors, making it a highly popular dataset to be used in mobile robotics and autonomous driving.
The TartanAir \cite{wang2020tartanair} is another benchmarking dataset for evaluation of \ac{SLAM} algorithms under challenging scenarios.
Additionally, the Imperial College London and National University of Ireland Maynooth (ICL-NUIM) \cite{handa2014benchmark} dataset is another \ac{VO} dataset containing handheld \ac{RGB-D} camera sequences, which has been used as a benchmark for many \ac{SLAM} works.

In contrast with the previous datasets, some other datasets contain data acquired using particular cameras instead of regular ones.
For instance, the \textit{Event Camera Dataset} introduced by Mueggler \etal \cite{mueggler2017event} is a dataset with collected samples using an event-based camera for high-speed robotic evaluations.
Dataset instances contain inertial measurements and intensity images captured by a motion-capture system, making it a suitable benchmark for \acp{VSLAM} equipped with event cameras.

\begin{figure*}[t!]
    \centering
    \caption{Instances of some of the most popular visual \ac{SLAM} datasets used for evaluation in various papers. The characteristics of these datasets can be found in Table~\ref{tbl_dataset}.}
    \label{fig:datasets}
    \includegraphics[width=0.9\linewidth]{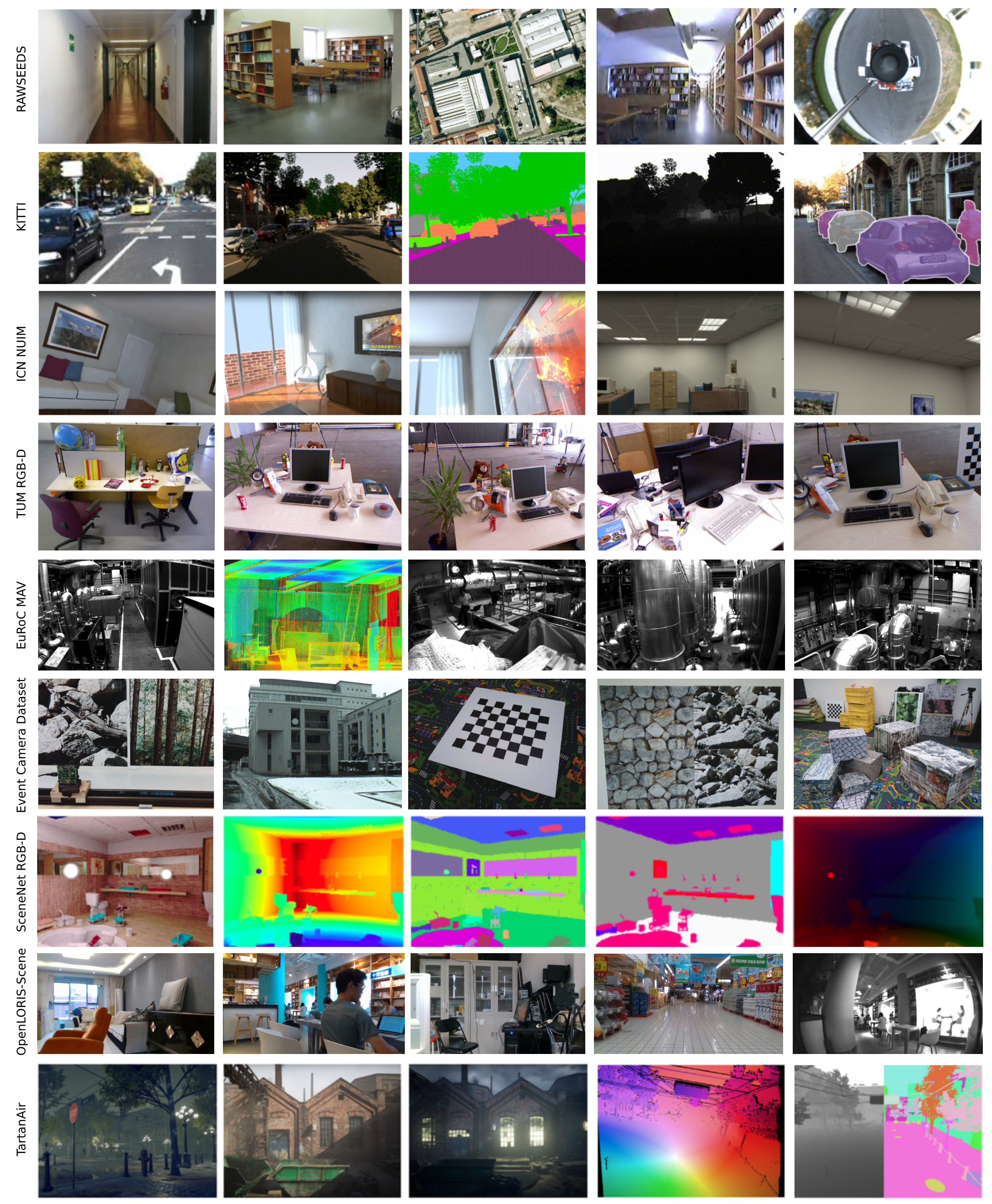}
\end{figure*}

The mentioned datasets are used in multiple \ac{VSLAM} methodologies according to their sensor setups, applications, and target environments.
These datasets mainly contain cameras' extrinsic and intrinsic calibration parameters and ground-truth data.
The summarized characteristics of the datasets and some instances of each are shown in Table~\ref{tbl_dataset} and \figurename{\ref{fig:datasets}}, respectively.

\begin{table*}[ht]
    \caption{Commonly used datasets for \ac{VSLAM} applications. \textit{GT} in the table refers to the availability of ground-truth values.}
        \label{tbl_dataset}
        \begin{tabular}{@{}lc|cc|cccccccccc|c@{}}
        \hline
        \multicolumn{1}{c}{\multirow{2}{*}{\textbf{Dataset Name}}} & \multirow{2}{*}{\textbf{Year}} & \multicolumn{2}{c}{\textbf{Environment}} \vline & \multicolumn{10}{c}{\textbf{Utilized Sensors}} \vline & \multicolumn{1}{c}{\multirow{2}{*}{\textbf{GT}}}
        \\
        \multicolumn{1}{c}{} & & \footnotesize{indoor} & \multicolumn{1}{c}{\footnotesize{outdoor}} \vline & \multicolumn{1}{c}{\footnotesize{GPS}} & \multicolumn{1}{c}{lidar} & \multicolumn{1}{c}{sonar} & \multicolumn{1}{c}{\ac{IMU}} & \multicolumn{1}{c}{mono} & \multicolumn{1}{c}{stereo} & \multicolumn{1}{c}{\ac{RGB-D}} & \multicolumn{1}{c}{event} & \multicolumn{1}{c}{omni} & \multicolumn{1}{c}{UWB} \vline & \multicolumn{1}{c}{} \\
        \hline
        RAWSEEDS \cite{bonarini2006rawseeds} & 2006 & \cmark & \cmark & \cmark & \cmark & \cmark & \cmark & & \cmark & & & \cmark & & \cmark \\
        KITTI \cite{geiger2012we} & 2012 & & \cmark & \cmark & \cmark & & \cmark & \cmark & \cmark & & & & & \cmark \\
        ICL-NUIM \cite{handa2014benchmark} & 2014 & \cmark & & & & & & & & \cmark & & & & \cmark \\
        TUM RGB-D \cite{sturm2012benchmark} & 2016 & \cmark & & & & & \cmark & & & \cmark & & & & \cmark \\
        EuRoC MAV \cite{burri2016euroc} & 2016 & \cmark & & & & & \cmark & \cmark & \cmark & & & & & \cmark \\
        Event Camera Dataset \cite{mueggler2017event} & 2017 & & \cmark & & & & \cmark & & & & \cmark & & & \cmark \\
        SceneNet RGB-D \cite{mccormac2017scenenet} & 2017 & \cmark & & & & & & & & \cmark & & & & \cmark \\
        OpenLORIS-Scene \cite{shi2020we} & 2020 & \cmark & & & \cmark & & \cmark & \cmark & \cmark & \cmark & & & & \cmark \\
        TartanAir \cite{wang2020tartanair} & 2020 & \cmark & \cmark & & \cmark & & & \cmark & \cmark & \cmark & & & & \cmark \\
        NTU VIRAL \cite{nguyen2021ntuviral} & 2021 & \cmark & \cmark & & \cmark & & \cmark & \cmark & & & & & \cmark & \cmark \\
        \hline
    \end{tabular}
\end{table*}

\subsection{Semantic Level} 
\label{sec:semantic}

Semantic information is required for the robot to understand the scene around it and make more profitable decisions.
In many recent \ac{VSLAM} works, adding semantic-level information to the geometry-based data is preferred to the pure geometry-based approaches, enabling them to deliver conceptual knowledge of the surroundings \cite{yu2018ds}.
In this regard, a pre-trained Object Recognition module can add semantic information to the \ac{VSLAM} models \cite{wen2021semantic}.
One of the most recent approaches is employing \acp{CNN} in \ac{VSLAM} applications.
In general, semantic \ac{VSLAM} approaches contain four primary components described below \cite{xu2019occupancy}:

\begin{itemize}
    \item \textit{Tracking module:} it uses the two-dimensional feature points extracted from consecutive video frames to estimate the camera pose and construct three-dimensional map points.
    Calculation of the camera pose and construction of the 3D map points build the baselines of the localization and mapping processes, respectively.
    \item \textit{Local mapping module}: by processing two sequential video frames, a new $3D$ map point is created, which is used along with a \ac{BA} module for an improved camera pose.
    \item \textit{Loop closing module}: by comparing the keyframes to the extracted visual features and assessing the similarities between them, it tunes the camera pose and optimizes the constructed map.
    \item \textit{\ac{NRCC}}: the main goal of employing \ac{NRCC} is to filter temporal objects from video frames in order to reduce their detrimental impact on the localization and mapping stages.
    It mainly contains a masking/segmentation process for separating various unstable instances in frames, such as people.
    Since it leads to lower the number of feature points to be processed, \ac{NRCC} simplifies the computational part and results in a more robust performance.
\end{itemize}

Accordingly, utilizing semantic level in \ac{VSLAM} approaches can improve the uncertainty in pose estimation and map reconstruction.
However, the current challenge here is to correctly use the extracted semantic information without hugely impacting the computational cost.
\section{VSLAM Approaches based on the Main Objectives}
\label{sec:focus}

In order to pinpoint \ac{VSLAM} approaches that achieve rich outcomes and present robust architectures, we collected and filtered out highly cited publications published in top-notch venues in the recent years from Google Scholar\footnote{\url{https://scholar.google.com/}} and well-known Computer Science bibliography databases: Scopus\footnote{\url{https://www.dblp.org/}} and DBLP\footnote{\url{https://www.scopus.com}}.
We also studied the manuscripts referred to in the mentioned publications and purified the ones most relevant to the \ac{VSLAM} domain.
After exploring the papers, we could categorize the collected publications based on their main objectives to solve a particular problems into sub-sections presented below:

\subsection{Objective I: Multi-sensor Processing}

This category covers the range of \ac{VSLAM} approaches that employ various sensors to understand the environment better.
While some techniques rely on \textit{only cameras} as the employed visual sensors, others combine \textit{various sensors} to enhance the accuracy of their algorithm.

\noindent \subsubsection{\textbf{Employing Multiple Cameras}} \hfill\\
As it might be difficult to recreate the 3D trajectories of moving objects with a single camera, some researchers suggest using multiple cameras instead.
For instance, \textit{CoSLAM\footnote{\url{https://github.com/danping/CoSLAM}}} is a \ac{VSLAM} system introduced by Zou and Tan \cite{zou2012coslam}, which uses separate cameras deployed on various platforms to reconstruct robust maps.
Their system combines multiple cameras moving around independently in a dynamic environment and reconstructs the map regarding their overlapping fields of view.
The process makes it easier to rebuild dynamic points in $3D$ by mixing intra- and inter-camera pose estimation and mapping.
\textit{CoSLAM} tracks visual features using the \ac{KLT} algorithm and operates in static and dynamic contexts, including indoors and outdoors, where the relative positions and orientations may shift over time.
The primary drawback of this method is requiring sophisticated hardware to interpret numerous camera outputs and increased computational cost by adding more cameras.

For challenging off-road settings, Yang \etal \cite{yang2020multi} developed a multi-camera cooperative panoramic vision \ac{VSLAM} approach.
Their approach gives each camera independence to increase the performance of the \ac{VSLAM} system under challenging conditions, such as occlusion and texture sparsity.
In order to determine the matching range, they extract \ac{ORB} features from cameras' overlapping fields of view.
Additionally, they employed a deep learning technique based on a \ac{CNN} to recognize similar features for loop closure detection.
For the experiment, the authors used a dataset produced by a panoramic camera and an integrated navigation system.

\textit{MultiCol-SLAM} is another an open-source \ac{VSLAM} framework with multi-camera configurations by Urban and Hinz \cite{urban2016multicol}.
They use their previously created model, \textit{MultiCol}, to enhance ORB-SLAM utilizing a keyframe-based process that supports multiple fisheye cameras.
They added a Multi-Keyframes (MKFs) processing module to ORB-SLAM, which collects turns images into keyframes.
Authors also proposed the idea of multi-camera loop closing, in which loop closures are detected from MKFs.
Although their method operates in real-time, it requires a significant computer power because several threads must run simultaneously.

\noindent \subsubsection{\textbf{Employing Multiple Sensors}} \hfill\\
Some other approaches proposed fusing various sensors and using vision- and inertial-based sensors outputs for better performance.
In this regard, a low-cost indirect lidar-assisted \ac{VSLAM} called \textit{CamVox\footnote{\url{https://github.com/ISEE-Technology/CamVox}}} was proposed by Zhu \etal \cite{zhu2021camvox} and demonstrated reliable performance and accuracy.
Their method uses ORB-SLAM 2.0 and combines the unique capabilities offered by Livox lidars as the premium depth sensors with the outputs from \ac{RGB-D} cameras.
The authors used an \ac{IMU} to synchronize and correct the non-repeating scanned locations.
Their contribution is presenting an autonomous lidar-camera calibration method that operates in uncontrolled environments.
Real-world tests on a robot platform indicate that \textit{CamVox} performs in real-time while processing the environment.

Authors in \cite{nguyen2021viral} proposed a multi-modal system titled \textit{VIRAL (Visual-Inertial-Ranging-Lidar) \ac{SLAM}} that couples camera, lidar, \ac{IMU}, and \ac{UWB}.
They also presented a map-matching marginalization scheme for visual features based on the local map constructed from lidar point clouds.
The visual components are extracted and tracked using \ac{BRIEF} algorithm.
The framework also contains a synchronization scheme and trigger for the utilized sensors.
They tested their approach on simulation environments and their generated dataset titled NTU VIRAL \cite{nguyen2021ntuviral} that contains data captured by camera, lidar, \ac{IMU}, and \ac{UWB} sensors.
However, their approach is computationally intensive due to handling synchronization, multi-threading, and sensor conflict resolution.

Vidal \etal \cite{vidal2018ultimate} proposed integrating events, camera frames, and \ac{IMU} in parallel configurations for reliable position estimation in high-speed settings.
Their \textit{Ultimate SLAM\footnote{\url{https://github.com/uzh-rpg/rpg_ultimate_slam_open}}} system is based on an event camera and a keyframe-based nonlinear optimization pipeline introduced in \cite{rebecq2017real}.
They use the \ac{FAST} corner detector and the Lucas-Kanade tracking algorithm for feature detection and tracking, respectively.
\textit{Ultimate SLAM} avoids motion blur problems brought on by high-speed activity and operates in dynamic situations with varied lighting conditions.
The efficiency of this technique on the "Event Camera Dataset" was obvious in comparison to alternative event-only and conventional camera configurations.
The authors also tested Ultimate SLAM on an autonomous quadrotor equipped with an event camera to show how their system can manage flight conditions that are impossible for conventional \ac{VO} platforms to handle.
The major challenge in \textit{Ultimate SLAM} is the synchronization of events with standard frame outputs.

A tightly-coupled monocular camera and \ac{UWB} range sensors were suggested by Nguyen \etal \cite{nguyen2020tightly} for \ac{VSLAM}.
They use a combination of feature-based (visible) and feature-less (\ac{UWB}) landmarks to create a map.
It operates effectively when \ac{UWB} is exposed to multi-path effects in congested surroundings.
They built their indirect method on ORB-SLAM and employ \ac{ORB} characteristics for pose estimation.
They tested their system on a generated dataset with hand-carried movements simulating an employed aerial robot.
The synchronization of the camera and \ac{UWB} sensor is one of the difficulties in this case, but it has been overcome by employing a new camera pose with its related timestamp for each new image.
\subsection{Objective II: Pose Estimation}

Methods classified in this category focus on how to improve the pose estimation of a \ac{VSLAM} approach using various algorithms.

\noindent \subsubsection{\textbf{Employing Lines/Points Data}} \hfill\\
In this regard, Zhou \etal \cite{zhou2015structslam} suggested employing building structural lines as useful features to determine the camera pose.
Structural lines are associated with dominant directions and encode global orientation information, resulting in improved predicted trajectories.
\textit{StructSLAM}, the mentioned method, is a 6-\ac{DoF} \ac{VSLAM} technique that operates in both low-feature and featureless conditions.
In employs \ac{EKF} to estimate variables based on the current directions in the scene.
For evaluation, the indoor scenes dataset from RAWSEEDS 2009 and a set of generated sequential images dataset were used.

\textit{Point and Line SLAM (PL-SLAM)}\footnote{\url{https://github.com/HarborC/PL-SLAM}}, a \ac{VSLAM} system based on ORB-SLAM optimized for non-dynamic low texture settings, was introduced by Pumarola \etal \cite{pumarola2017pl}.
The system simultaneously fuses line and point features for improved posture estimation and helps running in situations with few feature points.
The authors tested \textit{PL-SLAM} on their generated dataset and TUM RGB-D.
The drawback of their method is the computational cost and the essence of using other geometric primitives, \eg planes, for a more robust accuracy.

Gomez-Ojeda \etal \cite{gomez2019pl} introduced \textit{PL-SLAM\footnote{\url{https://github.com/rubengooj/pl-slam}}} (different from the framework with the same name by Pumarola \etal in \cite{pumarola2017pl}), an indirect \ac{VSLAM} technique that uses points and lines in stereo vision cameras to reconstruct an unseen map.
They merged segments obtained from points and lines in all \ac{VSLAM} modules with visual information taken from successive frames in their approach.
Using the \ac{ORB} and \ac{LSD} algorithms, points and line segments are retrieved and tracked in subsequent stereo frames in \textit{PL-SLAM}.
The authors tested \textit{PL-SLAM} on EuRoC and KITTI datasets and could outperform the stereo version of ORB-SLAM 2.0 in terms of performance.
One of the main drawbacks of \textit{PL-SLAM} is the computational time required for the feature tracking module and considering all structural lines to extract information about the environment.

A degeneracy avoidance technique for monocular point- and line-based \ac{VSLAM} systems was introduced by Lim \etal \cite{lim2021avoiding}.
A strong Optical Flow-based line tracking module that extracts line characteristics, filters out short lines in each frame, and matches the previously identified lines is another contribution of their method.
To demonstrate the efficacy of their technique and show that it was superior to the established point-based approaches, they tested their system on EuRoC MAV dataset.
The system lacks an adaptive approach to identify the correct optimization parameters, notwithstanding the strong findings.

\noindent \subsubsection{\textbf{Using Extra Features}} \hfill\\
\textit{Dual Quaternion Visual SLAM (DQV-SLAM)}, a framework for stereo-vision cameras that uses a broad Bayesian framework for 6-\ac{DoF} posture estimation was proposed in \cite{bultmann2019stereo}.
In order to prevent the linearization of the nonlinear spatial transformation group, their approach uses progressive Bayes updates.
For point clouds of maps and Optical Flow, \textit{DQV-SLAM uses} \ac{ORB} features to enable reliable data association in dynamic circumstances.
On KITTI and EuRoC datasets, the method could estimate experiment results reliably.
However, it lacks a probabilistic interpretation for stochastic modeling of poses and is computationally demanding for sampling approximation-based filtering.

Muñoz-Salinas \etal \cite{munoz2019spm} developed a technique using artificial squared planar markers to recreate a large-scale interior environment map.
Their real-time \textit{SPM-SLAM} system can solve the ambiguity issue of pose estimation using markers if at least two of them are visible in each video frame.
They created a dataset with video sequences of markers placed in two rooms joined by a door for examination.
Although \textit{SPM-SLAM} is cost-effective, it only works when numerous planar markers are scattered around the area while at least two are visible for marker connection recognition.
Moreover, the ability of their framework to handle dynamic changes in the scene is not measured.

\noindent \subsubsection{\textbf{Deep Learning}} \hfill\\
In another approach, Bruno and Colombini \cite{bruno2021lift} proposed \textit{LIFT-SLAM}, which combines deep learning-based feature descriptors with the conventional geometry-based systems.
They expanded the ORB-SLAM system's pipeline and employed a \ac{CNN} to extract features from images, using the learned features to provide more dense and precise matches.
For purposes of detection, description, and orientation estimation, \textit{LIFT-SLAM} fine-tunes a Learned Invariant Feature Transform (LIFT) deep neural network.
Studies using the KITTI and EuRoC MAV datasets' indoor and outdoor instances revealed that \textit{LIFT-SLAM} outperforms conventional feature-based and deep learning-based \ac{VSLAM} systems in terms of accuracy.
However, the weaknesses of the method are its computationally intensive pipeline and un-optimized \ac{CNN} design, which leads to near real-time performance.

Naveed \etal \cite{naveed2022deep} proposed a deep learning-based \ac{VSLAM} solution with a reliable and consistent module, even on routes with extreme turns.
Their approach outperformed several \acp{VSLAM} and used a deep reinforcement learning network trained on realistic simulators.
Furthermore, they provided a baseline for active \ac{VSLAM} evaluation and could properly generalize across actual indoor and outdoor environments.
The network's path planner developed the ideal path data, which is received by its base system, ORB-SLAM.
They produced a dataset with actual navigation episodes in challenging and texture-less environments for evaluation.

As another approach, \textit{RWT-SLAM} is a deep feature matching-based \ac{VSLAM} framework the authors in \cite{peng2022rwt} proposed for weakly textured situations.
Their method, which is based on ORB-SLAM, is fed with feature masks from an enhanced LoFTR \cite{sun2021loftr} algorithm for local image feature matching.
A \ac{CNN} architecture and the LoFTR algorithm were used to extract coarse-level and fine-level descriptors in the scene, respectively.
\textit{RWT-SLAM} is examined on the TUM RGB-D and OpenLORIS-Scene datasets, as well as a real-world dataset gathered by the authors.
However, their system is computationally demanding despite the robust feature matching results and performance.
\subsection{Objective III: Real-world Viability}

Approaches in this category have the primary objective of being used in various environments and working under several scenarios.
We notice that the references in this section are highly integrated with \textit{semantic} information extracted from the environment and present an end-to-end \ac{VSLAM} application.

\noindent \subsubsection{\textbf{Dynamic Environments}} \hfill\\
In this regard, a \ac{VSLAM} system titled \textit{DS-SLAM\footnote{\url{https://github.com/ivipsourcecode/DS-SLAM}}} has been introduced by Yu \etal \cite{yu2018ds}, which can be used in dynamic contexts and offers semantic-level information for map construction.
The system is built upon ORB-SLAM 2.0 and contains five threads: \textit{tracking}, \textit{semantic segmentation}, \textit{local mapping}, \textit{loop closing}, and \textit{dense semantic map construction}.
To exclude dynamic items before the pose estimation process and increase localization accuracy, \textit{DS-SLAM} employs the Optical Flow algorithm with a real-time semantic segmentation network called \textit{SegNet} \cite{badrinarayanan2017segnet}.
\textit{DS-SLAM} has been tested in real-world settings and with \ac{RGB-D} cameras, as well as on the TUM RGB-D dataset.
However, despite its high accuracy in localization, it faces semantically segmentation limitations and computationally intensive features.

\textit{Semantic Optical Flow SLAM (SOF-SLAM)}, an indirect \ac{VSLAM} system built upon the \ac{RGB-D} mode of ORB-SLAM 2.0, is another method in highly dynamic environments proposed by Cui and Ma \cite{cui2019sof}.
Their approach uses the Semantic Optical Flow dynamic feature detection module, which extracts and skips the changing features concealed in the semantic and geometric information provided by \ac{ORB} feature extraction.
In order to deliver accurate camera pose and environment reports, \textit{SOF-SLAM} makes use of SegNet's pixel-wise semantic segmentation module.
In extremely dynamic situations, experimental findings on the TUM RGB-D dataset and in real-world settings demonstrated that SOF-SLAM performs better than ORB-SLAM 2.0.
However, the ineffective method of non-static feature recognition and reliance on just two consecutive frames for this purpose are \textit{SOF-SLAM}'s weakness points.

Using the Optical Flow method to separate and eliminate dynamic feature points, Cheng \etal \cite{cheng2019improving} suggested a \ac{VSLAM} system for dynamic environments.
They have utilized the ORB-SLAM pipeline's structure and supplied it with fixed feature points generated from typical monocular camera outputs for precise posture estimation.
The system indicated operates in featureless circumstances by sorting Optical Flow values and using them for feature recognition.
According to experimental results on the TUM RGB-D dataset, the suggested system functions well in dynamic indoor circumstances.
However, the system's configuration uses an offline threshold for motion analysis, making it difficult to use in a variety of dynamic environment situations.

Another \ac{VSLAM} strategy was released by Yang \etal \cite{yang2022visual} that reconstructs the environment map using semantic segmentation network data, a motion consistency detection technique, and geometric restrictions.
Their approach, which is based on ORB-SLAM 2.0's \ac{RGB-D} variant, performs well in dynamic and indoor environments.
Only the stable features from the scene are retained using an improved \ac{ORB} feature extraction technique, while the dynamic characteristics are disregarded.
The features and the semantic data will then be combined to create a static semantic map.
Evaluation findings on the Oxford and TUM RGB-D datasets demonstrated the effectiveness of their approach in enhancing location accuracy and creating semantic maps with a wealth of data.
However, their system can run into problems in corridors or places with less information.

\noindent \subsubsection{\textbf{Deep Learning-based Solutions}} \hfill\\
In another work by Li \etal \cite{li2020dxslam} called \textit{DXSLAM\footnote{\url{https://github.com/ivipsourcecode/dxslam}}}, deep learning is used to find keypoints that resemble SuperPoints and to produce both the general descriptors and the images' keypoints.
They trained the cutting-edge deep \ac{CNN} HF-NET to produce frame- and keypoint-based descriptions by extracting local and global information from each frame.
They also used the offline \ac{BoW} method to train a visual vocabulary of local characteristics for precise loop closure recognition.
\textit{DXSLAM} operates in real-time without using a \ac{GPU} and is compatible with contemporary CPUs.
Even if such qualities are not specifically addressed, it has a great ability to resist dynamic changes in dynamic contexts.
\textit{DXSLAM} has been tested on TUM RGB-D and OpenLORIS-Scene datasets and both indoor and outdoor images and could achieve more accurate results than ORB-SLAM 2.0 and DS-SLAM.
However, the major disadvantages of this method are complex architecture for feature extraction and incorporating deep features into an old \ac{SLAM} framework.

In another approach, Li \etal \cite{li2021deep} developed a real-time \ac{VSLAM} technique for extracting feature points based on deep learning in complicated situations.
The method can run on a \ac{GPU} and supports the creation of 3D dense maps and is a multi-task \ac{CNN} for feature extraction with self-supervision capabilities.
The \ac{CNN} output is binary code strings with a fix-length of $256$, making it possible to be replaced by more conventional feature point detectors like \ac{ORB}.
It comprises three threads for reliable and timely performance in dynamic scenarios: \textit{tracking}, \textit{local mapping}, and \textit{loop closing}.
The system that supports monocular and \ac{RGB-D} cameras using ORB-SLAM 2.0 as a baseline.
The authors tested their methodology on the TUM dataset and two datasets collected in a corridor and an office using a Kinect camera for the experiments.

Steenbeek and Nex in \cite{steenbeek2022cnn} a real-time \ac{VSLAM} technique that uses a \ac{CNN} for accurate scene interpretation and map reconstruction.
Their solution utilizes monocular camera streams from a \ac{UAV} during flight and employs a depth-estimating neural network for reliable performance.
The mentioned method is based on ORB-SLAM 2.0 and makes use of visual cues collected from indoor environments.
Additionally, the \ac{CNN} is trained on more than 48,000 indoor examples and operates the pose, space depth, and RGB inputs to estimate scale and depth.
The TUM RGB-D dataset and a real-world test using a drone were used to evaluate the system, which demonstrated enhanced pose estimation accuracy.
However, the system struggles in situations without texture and needs both CPU and GPU resources for real-time performance.

\noindent \subsubsection{\textbf{Using Artificial Landmarks}} \hfill\\
A technique called \textit{UcoSLAM\footnote{\url{https://sourceforge.net/projects/ucoslam/}}} \cite{munoz2020ucoslam} by Muñoz-Salinas and Medina-Carnicer outperforms conventional \ac{VSLAM} systems by combining natural and man-made landmarks and automatically calculating the scale of the surroundings using fiducial markers.
\textit{UcoSLAM}'s primary driving force is to combat natural landmarks' instability, repetition, and poor tracking qualities.
It can operate in surroundings without tags or features since it can operate in keypoints-only, markers-only, and mixed modes.
To locate map correspondences, optimize re-projection errors, and re-localize in the event of tracking failure, \textit{UcoSLAM} has a tracking mode.
Additionally, it has a marker-based loop closure detection system and can describe features using any descriptor, including \ac{ORB} and \ac{FAST}.
Despite all the plus points of \textit{UcoSLAM}, the system executes in multiple threads, making it a time-consuming approach.

\noindent \subsubsection{\textbf{Wide-range of Setups}} \hfill\\
Another \ac{VSLAM} strategy for dynamic indoor and outdoor situations is \textit{DMS-SLAM} \cite{liu2019dms}, which supports monocular, stereo, and \ac{RGB-D} visual sensors.
The system employs sliding window and \ac{GMS} \cite{bian2017gms} feature matching methods to find static feature locations.
Using the ORB-SLAM 2.0 system as its foundation, \textit{DMS-SLAM} tracks the static features recognized by the \ac{ORB} algorithm.
The authors tested their suggested methodology on the TUM \ac{RGB-D} and KITTI datasets and outperformed cutting-edge \ac{VSLAM} algorithms.
Additionally, because the feature points on the dynamic objects were removed during the tracking step, \textit{DMS-SLAM} performs more quickly than the original ORB-SLAM 2.0.
Despite the benefits described, the suggested solution encounters difficulties in situations with little texture, fast motion, and highly dynamic environments.
\subsection{Objective IV: Resource Constraint}

In another category, some of the \ac{VSLAM} methodologies are built for devices with limited computational resources compared to other standard devices.
For instance, \ac{VSLAM} systems designed for mobile devices and robots with embedded systems are included in this category.

\noindent \subsubsection{\textbf{Devices with Limited Processing Capabilities}} \hfill\\
In this regard, \textit{edgeSLAM} is a real-time, edge-assisted semantic \ac{VSLAM} system for mobile and resource-constrained devices proposed by Xu \etal \cite{xu2020edge}.
It employs a series of fine-grained modules to be used by an edge server and the associated mobile devices rather than requiring heavy threads.
A semantic segmentation module based on the Mask-RCNN technique is also included in \textit{edgeSLAM} to improve segmentation and object tracking.
The authors put their strategy into practice on an edge server with several commercial mobile devices, such as cellphones and development boards.
By reusing the findings of the object segmentation, they avoided duplicate processing by adapting system parameters to different network bandwidth and latency situations.
\textit{EdgeSLAM} has been evaluated on monocular vision instances of TUM RGB-D, KITTI, and the created dataset for experimental settings.

For stereo camera setups, Schlegel, Colosi, and Grisetti \cite{schlegel2018proslam} suggested a lightweight feature-based \ac{VSLAM} framework titled \textit{ProSLAM\footnote{\url{https://gitlab.com/srrg-software/srrg_proslam}}} that achieves results on par with cutting-edge techniques.
Four modules make up their approach: the \textit{triangulation} module, which creates 3D points and associated feature descriptors; the \textit{incremental motion estimation} module, which processes two frames to determine the current position; the \textit{map management} module, which creates local maps; and the \textit{re-localization} module, which updates the world map based on the similarities of local maps.
\textit{ProSLAM} retrieves the 3D position of the points using a single thread and leverages a small number of well-known libraries for a system that is simple to create.
According to the experiments on KITTI and EuRoC datasets, their approach could achieve robust results.
However, it shows weakness in rotation estimation and does not contain any bundle adjustment module.

Bavle \etal \cite{bavle2020vps} proposed \textit{VPS-SLAM~\footnote{\url{https://github.com/hridaybavle/semantic_slam}}}, a lightweight graph-based \ac{VSLAM} framework for aerial robotics.
Their real-time system integrates geometrical data, several object detection techniques, and visual/visual-inertial odometry for pose estimation and building the semantic map of the environment.
Low-level characteristics, \ac{IMU} measurements, and high-level planar information are all used by VPS-SLAM to reconstruct sparse semantic maps and predict robot states.
The system leverages the lightweight version of \ac{YOLO2} \cite{redmon2017yolo9000} trained on the COCO dataset \cite{lin2014microsoft} for object detection due to its real-time and computationally effective performance.
They used a hand-held camera setup and an aerial robotic platform equipped with an \ac{RGB-D} camera for testing.
The TUM RGB-D dataset's indoor instances were used to test their methodology, and they were able to provide results that were on par with those of well-known \ac{VSLAM} methods.
However, only a small number of objects (\eg chairs, books, and laptops) can be used by their \ac{VSLAM} system to build a semantic map of the surrounding area.

Another real-time indoor \ac{VSLAM} method was proposed by Tseng \etal \cite{tseng2022real} that requires a low-cost setup.
The authors also presented a technique for estimating the number of frames and visual elements required for a reasonable degree of localization accuracy.
Their solution is based on the OpenVSLAM \cite{sumikura2019openvslam} framework and makes use of it for emergencies that arise in the real world, such as gaining access to specific targets.
The system acquires the scene's feature map for precise pose estimation by applying the \ac{EPnP} and \ac{RANSAC} algorithms.
According to tests conducted in a building, their device can deliver accurate findings under difficult lighting conditions.

\noindent \subsubsection{\textbf{Computation Offloading}} \hfill\\
Ben Ali \etal \cite{ben2020edge} suggested using edge computing to enable the offloading of resource-intensive operations to the cloud and reduce the computational burden on the robot.
They modified the architecture of ORB-SLAM 2.0 in their indirect framework, Edge-SLAM\footnote{\url{https://github.com/droneslab/edgeslam}}, by maintaining the tracking module on the robot and delegating the remainder to the edge.
By splitting the \ac{VSLAM} pipeline between the robot and the edge device, the system can maintain both a local and a global map.
With fewer resources available, they could still execute properly without sacrificing accuracy.
They used the TUM RGB-D dataset and two different mobile devices to generate a custom indoor environment dataset using RGB-D cameras for evaluation.
However, one of their approach's drawbacks is the architecture's complexity due to the decoupling of various \ac{SLAM} modules.
Another setback is that their system works only in short-term settings, and utilizing Edge-SLAM in long-term scenarios (\eg multiple days) would face performance degradation.
\subsection{Objective V: Versatility}

\ac{VSLAM} works categorized in this class are focused on straightforward development, utilization, adaptation, and extension.

In this regard, Sumikura \etal \cite{sumikura2019openvslam} introduced \textit{OpenVSLAM\footnote{\url{https://github.com/xdspacelab/openvslam}}}, a highly adaptable open-source \ac{VSLAM} framework seeks to be quickly developed upon and called by other third-party programs.
Their feature-based approach is compatible with multiple camera types, including monocular, stereo, and \ac{RGB-D}, and can store or reuse the reconstructed maps for later usage.
\textit{OpenVSLAM} performs better in terms of tracking accuracy and efficiency than ORB-SLAM and ORB-SLAM 2.0 due to its powerful \ac{ORB} feature extractor module.
However, the open-source code of the system has been discontinued owing to worries over code similarities that infringed on the rights to ORB-SLAM 2.0.

To bridge the gap between real-time capabilities, accuracy, and resilience, Ferrera \etal \cite{ferrera2021ov} developed \textit{{OV$^{2}$SLAM}\footnote{\url{https://github.com/ov2slam/ov2slam}}} that works with monocular and stereo-vision cameras.
By limiting the extraction of features to keyframes and monitoring them in subsequent frames through eliminating photo-metric errors, their method lessens the computational load.
In this sense, \textit{{OV$^{2}$SLAM}} is a hybrid strategy that combines the virtues of the direct and indirect categories of \ac{VSLAM} algorithms.
Using well-known benchmarking datasets including EuRoC, KITTI, and TartanAir in both indoor and outdoor experiments, it was demonstrated that \textit{{OV$^{2}$SLAM}} surpasses several popular techniques in terms of performance and accuracy.

Another approach in this category, titled \textit{DROID-SLAM\footnote{\url{https://github.com/princeton-vl/DROID-SLAM}}}, a deep learning-based visual \ac{SLAM} for monocular, stereo, and \ac{RGB-D} cameras, is proposed by Teed and Deng \cite{teed2021droid}.
They could attain greater accuracy and robustness than well-known monocular and stereo track methods.
Their solution operates in real-time and consists of \textit{back-end} (for bundle adjustment) and \textit{front-end} (for keyframe collection and graph optimization) threads.
\textit{DROID-SLAM} has already been taught using monocular camera examples, therefore it does not need to be trained again to use stereo and \ac{RGB-D} inputs.
The approach minimizes the projection error, like indirect methods, while not requiring any pre-processing for feature identification and matching.
A feature extraction network comprising downsampling layers and residual blocks processes each input image to create dense features.
\textit{DROID-SLAM} has been tested on well-known datasets, including TartanAir, EuRoC, and TUM RGB-D, and could achieve acceptable results.

Bonetto \etal in \cite{bonetto2022irotate} propose \textit{iRotate~\footnote{\url{https://github.com/eliabntt/irotate_active_slam}}}, an active technique for omnidirectional robots with \ac{RGB-D} cameras.
Additionally, a module for spotting obstructions in the camera's area of vision is employed in their approach.
By offering observation coverage of previously unexplored places and previously visited locations, \textit{iRotate}'s primary objective is to lessen the distance the robot must go to map the environment.
The mentioned method uses a \ac{VSLAM} framework with graph features as its back-end.
The authors could attain outcomes that were on par with those of cutting-edge \ac{VSLAM} methods by providing comparisons in simulation and on a real three-wheel omnidirectional robot.
However, the major weakness of their method is that the robot might face start-stop cases in which the local paths are re-planned.
\subsection{Objective VI: Visual Odometry}

Approaches in this category aim to determine the position and orientation of the robot with the highest possible accuracy.

\noindent \subsubsection{\textbf{Deep Neural Networks}} \hfill\\
In this regard, the \textit{Dynamic-SLAM} framework was proposed in \cite{xiao2019dynamic} that leverages deep learning for accurate pose prediction and suitable environment comprehension.
As part of a semantic level module for optimized \ac{VO}, the authors employed a \ac{CNN} to identify moving objects in the environment, which helped them lower the pose estimate error brought on by improper feature matching.
Additionally, \textit{Dynamic-SLAM} uses a selective tracking module to ignore dynamic locations in the scene and a missed feature corrective algorithm for speed invariance in adjacent frames.
Despite the excellent results, the system requires huge computational costs and faces the risk of misclassifying dynamic/static objects due to a limited number of defined semantic classes.

Bloesch \etal \cite{bloesch2018codeslam} proposed the \textit{Code-SLAM\footnote{\url{https://github.com/silviutroscot/CodeSLAM}}} direct technique, which offers a condensed and dense representation of the scene geometry.
Their \ac{VSLAM} system, which only functions with monocular cameras, is an enhanced version of \textit{PTAM} \cite{klein2007parallel}.
They divided intensity images into convolutional features and fed them to a depth auto-encoder using a \ac{CNN} trained on intensity images from the SceneNet RGB-D dataset.
Indoor examples of the EuRoC dataset's have been used to test \textit{Code-SLAM}, and the findings were promising in terms of accuracy and performance.

\textit{DeepVO\footnote{\url{http://senwang.gitlab.io/DeepVO/}}}, an end-to-end \ac{VO} framework using a Deep \ac{RCNN} architecture for monocular settings, was proposed by Wang \etal \cite{wang2017deepvo}.
Their approach uses deep learning to automatically learn the appropriate features, model sequential dynamics and relations, and infer poses directly from color frames.
The \textit{DeepVO} architecture includes a \ac{CNN} called FlowNet for computing optical flow from sequential frames and two \ac{LSTM} layers for estimating the temporal changes based on the feed provided by the \ac{CNN}.
The framework can simultaneously extract visual characteristics and perform sequential modeling by combining \ac{CNN} and \ac{RNN}.
\textit{DeepVO} can incorporate geometry with the knowledge models learned for an enhanced \ac{VO}.
However, it cannot be utilized to replace conventional geometry-based \ac{VO} approaches.

%
Parisotto \etal \cite{parisotto2018global} proposed a DeepVO-like end-to-end system using a Neural Graph Optimization (NGO) step instead of \acp{LSTM}.
Their approach operates a loop closure detection and correction mechanism based on temporally distinct poses.
NGO uses two attention-optimization methods to jointly optimize the aggregated predictions made by convolutional layers of local pose estimation modules and delivers global pose estimations.
They experimented with their technique on $2D$ and $3D$ mazes and outperformed DeepVO's performance and accuracy levels.
The mentioned approach needs to be connected to a \ac{SLAM} framework to supply the re-localization signals.

In another work, one of the most extensive \ac{VSLAM} frameworks titled \textit{DeepFactors\footnote{\url{https://github.com/jczarnowski/DeepFactors}}} introduced by Czarnowski \etal \cite{czarnowski2020deepfactors} for densely rebuilding the environment map with monocular cameras.
For a more reliable map reconstruction, their real-time solution performs joint optimization of the pose and depth, makes use of probabilistic data, and combines learned and model-based approaches.
The authors modified the \textit{CodeSLAM} framework and added missing components such as local/global loop detection.
The system is evaluated on the ICL-NUIM and TUM RGB-D datasets after being trained on roughly 1.4 million ScanNet \cite{dai2017scannet} images.
\textit{DeepFactors} improves the idea of the CodeSLAM framework and focuses on code optimization in traditional \ac{SLAM} pipelines.
However, due to the computational costs of the modules, this approach requires employing \acp{GPU} to guarantee real-time performance.

\noindent \subsubsection{\textbf{In-depth Adjacent Frame Processing}} \hfill\\
In another work, and by reducing the photometric and geometric errors between two pictures for camera motion detection, the authors of \cite{kerl2013dense}\footnote{\url{https://vision.in.tum.de/data/software/dvo}} developed a real-time dense \ac{SLAM} approach for \ac{RGB-D} cameras, improving their prior method, \cite{kerl2013robust}.
Their keyframe-based solution expands \textit{Pose SLAM} \cite{ila2009information} that only keeps non-redundant poses for producing a compact map, adds dense visual odometry characteristics, and effectively utilizes the information from camera frames for a reliable camera motion estimation.
Authors also employed an entropy-based technique to gauge keyframe similarity for loop closure detection and drift avoidance.
However, their approach still needs work in the areas of loop closure detection and keyframe selection quality.

In another work introduced by Li \etal \cite{li2021dp}, real-time dynamic object removal is accomplished using a feature-based \ac{VSLAM} approach known as \textit{DP-SLAM}.
This method uses a Bayesian probability propagation model based on the likelihood of the keypoints derived from moving objects.
The variation of geometry restrictions and semantic data can be overcome by \textit{DP-SLAM} using the moving probability propagation algorithm and iterative probability updates.
It is integrated with ORB-SLAM 2.0 and has been tested on the TUM RGB-D dataset.
Despite the accurate results, the system only works in sparse \acp{VSLAM} and faces high computational costs because of the iterative probability updater module.

\textit{Pair-Navi}, a suggested indoor navigation system by Dong \etal \cite{dong2019pair}, reuses a previously traced path by an agent for usage in the future by other agents.
Hence, a previous traveler known as the \textit{leader} captures the trace information, such as turnings and particular ambient qualities, and gives it to a later \textit{follower} that needs to travel to the same destination.
While the \textit{follower} uses a re-localization module to determine its location concerning the reference trace, the \textit{leader} incorporates visual odometry and trajectory creation modules.
To recognize and remove dynamic items from the video feature set, the system employs a Mask Region-based \ac{CNN} (Mask R-CNN).
They tested \textit{Pair-Navi} on a set of generated datasets and several smartphones for the experiments.

\noindent \subsubsection{\textbf{Various Feature Processing}} \hfill\\
Another approach in this category is a text-based \ac{VSLAM} system called \textit{TextSLAM}, proposed by Li \etal \cite{li2020textslam}.
It incorporates text items retrieved from the scene using the \ac{FAST} corner detection technique into the \ac{SLAM} pipeline.
Texts include a variety of textures, patterns, and semantic meanings, making the approach more efficient to use them to create 3D text maps of high quality.
\textit{TextSLAM} uses texts as reliable visual fiducial markers, parametrizes them after the first frame in which they are found, and then projects the 3D text object onto the target image to locate it again.
They also presented a new three-variable parameterization technique for initializing instantaneous text features.
Using a monocular camera and a dataset created by the authors, experiments were conducted in indoor and outdoor settings, and the results were highly accurate.
Operating in text-free surroundings, interpreting short letters, and requiring the storage of enormous text dictionaries are the three fundamental challenges of \textit{TextSLAM}.

Xu \etal \cite{xu2019occupancy} proposed an indirect \ac{VSLAM} system built upon a modified ORB-SLAM that enables high-accuracy localization and user interaction using the \ac{OGM} method and a new 2D mapping module.
Their system can reconstruct the environment map that shows the presence of a barrier as an equally spaced field of variables using the \ac{OGM}, makes it possible to navigate continuously and in real-time while planning a route.
The experimental examination of a generated dataset shows their approach functions in GPS-denied conditions.
However, their technique struggles to function well in dynamic and complicated environments and has trouble appropriately matching the features in corridors and featureless conditions.

\textit{CPA-SLAM}, a direct \ac{VSLAM} method for \ac{RGB-D} cameras that makes use of planes for tracking and graph optimization, was proposed by Ma \etal \cite{ma2016cpa}.
\textit{Frame-to-keyframe} and \textit{frame-to-plane} alignments are regularly integrated in their technique.
They also introduced an image alignment algorithm for tracking the camera with respect to a reference keyframe and aligning the image with the planes.
The keyframe data is used by \textit{CPA-SLAM}, which looks for the closest short temporal and geographical distances to track.
The system's real-time performance of their tracking system was tested in with- and without-plane settings, with analyses performed on TUM RGB-D and ICL-NUIM datasets with indoor and outdoor scenes.
However, it only supports a small number of geometric shapes, \ie planes.
\section{Pinpointing the Current Trends}
\label{sec:discussion}

\subsection{Statistics}

\begin{figure*}[t!]
    \centering
    \caption{Analyzing the current trends of \ac{VSLAM} approaches: a) base \ac{SLAM} system employed to implement a new approach, b) the primary objective of the approach, c) various datasets that the proposed methods were testing on, d) the impact of utilizing semantic data in the proposed methods, e) the amount of dynamic objects existing in the environment, f) various types of environments the system tested on.}
    \label{fig:discussion}
    \includegraphics[width=0.9\linewidth]{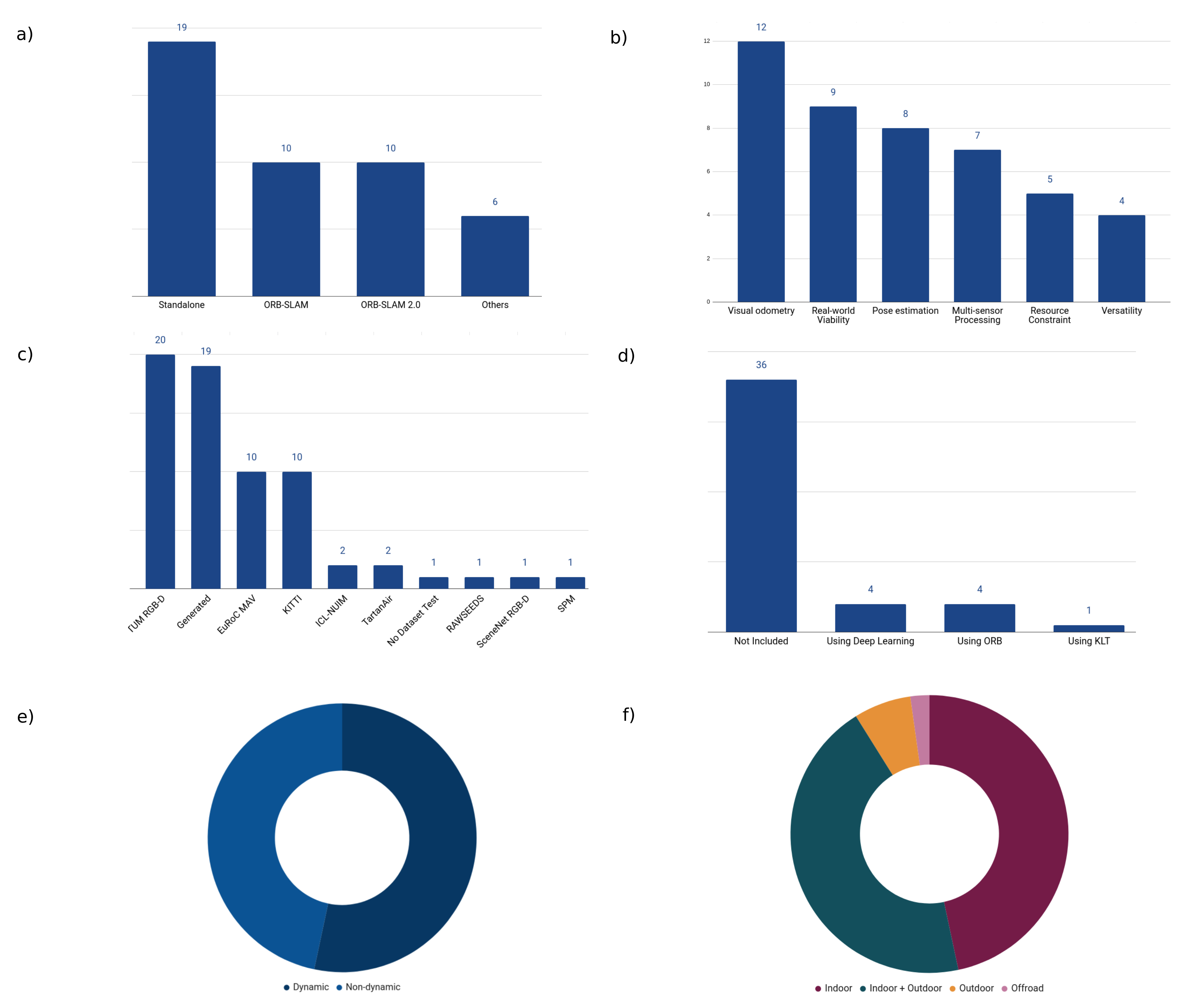}
\end{figure*}

Regarding the classification of the surveyed papers in various aspects presented above, we have visualized the processed data in \figurename{\ref{fig:discussion}} to find the current trends in \ac{VSLAM}.
In sub-figure \textit{"a"}, We can see that the majority of the proposed \ac{VSLAM} systems are standalone applications that implement the whole procedure of localization and mapping using visual sensors from scratch.
While ORB-SLAM 2.0 and ORB-SLAM are other \textit{base} platforms employed to make a new framework, minimal approaches are based on other \ac{VSLAM} systems, such as \ac{PTAM} and PoseSLAM.
Moreover, and in terms of the objectives of \ac{VSLAM} applications, what tops the chart in sub-figure \textit{"b"} is improving the \textit{\acl{VO}} module.
Thus, most of the recent \acp{VSLAM} are trying to resolve the problems of current algorithms in determining the position and orientation of the robots.
\textit{Pose estimation} and \textit{Real-world viability} are further fundamental objectives of proposing new \ac{VSLAM} papers.
Concerning the dataset used for evaluation in the surveyed papers, sub-figure \textit{"c"} illustrates that most works have been tested on the \textit{TUM RGB-D} dataset.
This dataset has been employed as the primary or one of the multiple baselines for evaluation in the reviewed manuscripts.
Additionally, many researchers tend to perform experiments on the datasets \textit{generated} by them.
We can assume that the primary motivation for generating a dataset was to show how a \ac{VSLAM} method works in real-world scenarios and if it can be used as an end-to-end application.
\textit{EuRoC MAV} and \textit{KITTI} are the next popular datasets for evaluation in \ac{VSLAM} works, respectively.
Another interesting information extracted from sub-figure \textit{"d"} is concerning the impact of employing semantic data when using the \ac{VSLAM} system.
We can see that the majority of the reviewed papers do not include semantic data while processing the environment.
We presume the reason behind not utilizing the semantic data are:
\begin{itemize}
    \item The computational cost for training a model that recognizes objects and utilizing it for semantic segmentation is considerable in many cases, which might raise the processing time.
    \item The majority of the geometry-based \ac{VSLAM} works are designed to work as plug-and-play devices so that they can employ camera data for localization and mapping with the least possible effort.
    \item Incorrect information extracted from the scene can also lead to more added noise to the process.
\end{itemize}
When the environment is considered, we can see in sub-figures \textit{"e"} that more that half of the approaches can also work in dynamic environments with challenging conditions, while the remaining systems are only focused on environment with no dynamic changes.
Moreover, and in sub-figure \textit{"f"}, most of the approaches work in "indoor settings" or "both indoor and outdoor environments," while the rest of the papers have only been tested outdoor conditions.
It should be mentioned that approaches that can only work in a particular circumstance with restricted assumptions might not produce the same accuracy if employed in other scenarios.
That is one of the main reasons why some approaches only concentrate on a particular situation.

\subsection{Analyzing the Current Trends}

The current survey has reviewed the state-of-the-art visual \ac{SLAM} approaches that have absorbed massive attention and demonstrated their principal contributions in this field.
Despite the wide range of reliable solutions and improvements in various modules of \ac{VSLAM} systems over the past years, there are still many high potential fields and unsolved issues that investigating in them lead to more robust approaches in the future evolution of \acp{SLAM}.
In light of the wide range of visual \ac{SLAM} approaches, we hereby propose currently trending areas for investment and introduce the following open research directions:

\noindent \textbf{Deep Learning}:
deep neural networks have shown encouraging results in various applications, including \ac{VSLAM} \cite{duan2020deep}, making them a significant trend in multiple fields of study.
Due to their learning capabilities, these architectures have shown a considerable potential to be utilized as reliable feature extractors to tackle different issues in \ac{VO} and loop closure detection.
\acp{CNN} can aid \acp{VSLAM} in precise object detection and semantic segmentation, and can outperform traditional feature extraction and matching algorithms for correctly recognizing hand-crafted features.
It has to be mentioned that since deep learning-based methods have been trained on datasets with large amounts of diversified data and limited object classes, there is always a risk of misclassification of dynamic points and causing false segmentation.
Thus, it might lead to lower segmentation accuracy and pose estimation error.

\noindent \textbf{Information Retrieval and Computational Cost Trade-off}:
generally, the processing cost and the quantity of information in the scene should always be balanced.
In this perspective, dense maps allow \ac{VSLAM} applications to record high-dimensional complete scene information, but doing so in real-time would be computationally demanding.
Sparse representations, on the other hand, would fail to capture all the needed information due to their lower computational cost.
It should also be noted that real-time performance is directly related to the camera’s frame rate, and frame losses in peak processing times can negatively affect the \ac{VSLAM} system’s performance, regardless of algorithms performance.
Moreover, \acp{VSLAM} typically take advantage of tightly-coupled modules and modifying one module may adversely affect others, which makes the balancing task more challenging.

\noindent \textbf{Semantic Segmentation}:
Providing semantic information while creating the map of the environment can bring about very useful information for the robot.
Identifying objects in the camera's field of view \textit{-\eg doors, windows, people, \etc-} are a trendy topic in current and future \ac{VSLAM} works, as the semantic information can be used in pose estimation, trajectory planning, and loop closure detection modules.
With the widespread usage of object detection and tracking algorithms, semantic \acp{VSLAM} will undoubtedly be among the future solutions in this domain.

\noindent \textbf{Loop Closing Algorithms}:
One of the critical issues in any \ac{SLAM} system is the drift problem and losing the feature tracks caused by accumulated localization errors.
Detection of drifts and loop closures to identify previously visited places contributes to high computation latency and cost in \ac{VSLAM} systems \cite{xu2020edge}.
The main reason is that the complexity of loop closure detection increases with the size of the reconstructed map.
Moreover, combining the map data collected from various locations and refining the estimated poses are very complex tasks.
With this, optimization and balancing of the loop closure detection module have a massive potential for improvement.
One of the common approaches for detecting loop closures is improving image retrieval by training a visual vocabulary based on local features and then aggregating them.

\noindent \textbf{Working in Challenging Scenarios}:
Working in a texture-less environment with few salient feature points often leads to drift errors in position and orientation in robots.
As one of the primary challenges in \ac{VSLAM}, this error may lead to system failure.
Thus, considering complementary scene-understanding methods in feature-based approaches, such as object detection or line features, would be a trendy topic.
\section{Conclusions}
\label{sec:conclusion}

This paper presented a broad range of \ac{SLAM} works in which visual data collected from cameras play a significant role.
We categorized the recent works of \ac{VSLAM} systems based on various characteristics of their approaches, such as the experimental environment, novelty domain, object detection and tracking algorithms, semantic level viability, performance, \etc
We also reviewed the critical contributions of the works and the existing drawbacks and challenges according to the authors' claims, future version improvements, and the issues addressed in other related methods.
Another contribution of the paper is discussing the current trends of \ac{VSLAM} systems and the existing open issues that will be investigated more by researchers.

\bibliographystyle{IEEEtran}
\bibliography{mybib}

\end{document}